\documentclass{article}

\usepackage{graphicx}%
\usepackage{multirow}%
\usepackage{amsmath,amssymb,amsfonts}%
\usepackage{amsthm}%
\usepackage{mathrsfs}%
\usepackage[title]{appendix}%
\usepackage{xcolor}%
\usepackage{textcomp}%
\usepackage{manyfoot}%
\usepackage{booktabs}%
\usepackage{algorithm}%
\usepackage{algorithmicx}%
\usepackage{algpseudocode}%
\usepackage{listings}%
\usepackage{caption}
\usepackage{multicol}
\usepackage{bm}
\usepackage{todonotes}
\usepackage{changepage}
\usepackage[marginclue,footnote,silent,draft]{fixme}
\usepackage{url}

\graphicspath{ {images/} }

\begin{document}

\title{Optimal Density Functions for Weighted Convolution in Learning Models}


\makeatletter
\onecolumn
{\fontsize{18pt}{20pt}\selectfont\bfseries\@title\par}
Simone Cammarasana
 \footnote{
\textbf{Simone Cammarasana}
CNR-IMATI, Via De Marini 6, Genova, Italy \\
simone.cammarasana@ge.imati.cnr.it
}
  , 
 Giuseppe Patan\`e
  \footnote{
 \textbf{Giuseppe Patan\`e} 
 CNR-IMATI, Via De Marini 6, Genova, Italy 
}

\makeatother


\abstract{The paper introduces the \emph{weighted convolution}, a novel approach to the convolution for signals defined on regular grids (e.g., 2D images) through the application of an optimal density function to scale the contribution of neighbouring pixels based on their distance from the central pixel. This choice differs from the traditional uniform convolution, which treats all neighbouring pixels equally. Our weighted convolution can be applied to convolutional neural network problems to improve the approximation accuracy. Given a convolutional network, we define a framework to compute the optimal density function through a minimisation model. The framework separates the optimisation of the convolutional kernel weights (using stochastic gradient descent) from the optimisation of the density function (using DIRECT-L). Experimental results on a learning model for an image-to-image task (e.g., image denoising) show that the weighted convolution significantly reduces the loss (up to~$53\%$ improvement) and increases the test accuracy compared to standard convolution. While this method increases execution time by~$11\%$, it is robust across several hyperparameters of the learning model. Future work will apply the weighted convolution to real-case 2D and 3D image convolutional learning problems.}
\textbf{Keywords:} Weighted convolution, Optimal density function, Optimisation model, Deep learning


\section{Introduction}\label{sec:INTRO}
Deep learning is a class of artificial intelligence methods for the processing and analysis of signals defined on regular grids (e.g., 2D images). Deep learning utilises large data sets and multiple levels of representation, comprising several linear and non-linear layers that transform the input signal into a higher-level representation, thereby reducing the number of parameters that represent the signal. Deep learning is widespread for the solution of complex problems in several image-based applications, such as computer vision~\cite{voulodimos2018deep}, robotics~\cite{pierson2017deep}, automotive~\cite{luckow2016deep}, and healthcare~\cite{cammarasana2022real}.

The convolution operator is applied to 2D images for its property of feature extraction and description of spatial hierarchies and local dependencies. The application of the convolution into deep learning leads to a sub-class of models named \emph{convolutional neural networks} (CNNs). CNNs apply the convolution filters across the input images, accounting for the information redundancy and enhancing the detection of local patterns and image features. This approach reduces the number of trainable parameters, thereby decreasing memory usage and computational cost, while generalising across different and large datasets. In CNNs, the convolution equally scales the neighbourhood elements of the reference pixel, while the weights of the kernel are optimised to improve the extraction of features from the dataset and reduce the loss function. Within this approach, the contribution of the pixels in the neighbourhood of the reference one depends only on the trainable weights (i.e., the features to be extracted), assuming that all the pixels have the same relevance~(Sect.~\ref{sec:BGRW}).

We introduce our \emph{weighted convolution} as the application of a density function to the convolution operator in CNNs to account for the position of the pixels with respect to the reference one. In particular, we define an optimisation model that computes the optimal density function to be applied to the convolution, i.e., the scaling of the neighbourhood of each pixel that minimises the loss function with respect to the uniform scaling, given the same number of trainable weights~(Sect.~\ref{sec:METHOD}). We focus our model on 2D images; however, our framework is general enough to be applied to any signal defined on 3D images.
 
We define a learning model with an image-to-image task (e.g., image denoising) and a learning architecture with convolutional layers and non-linear functions. In this learning model, a density function is applied to the convolution operator. The learning model is trained to optimise the weights of the kernel to minimise a loss function, i.e., the distance between the target and predicted image. The learning model is the objective function of our optimisation model, where the density function values are the variables to be optimised. This approach allows us to separate the optimisation of the learning model (i.e., the kernel weights) from the optimisation of the density function. We apply two different optimisers for the two learning problems. The \emph{stochastic gradient descent} (SGD)~\cite{amari1993backpropagation} is applied to minimise the loss function of the learning problem. The SGD is a local method that applies to differentiable functions by updating parameters in the direction of the negative gradient. The \emph{divide rectangle-local} (DIRECT-L)~\cite{gablonsky2000locally} is applied to compute the optimal density function. DIRECT-L is a global optimisation method that is used for optimising non-differentiable functions. We also underline that our goal is not the implementation of an efficient learning architecture, but the computation of the optimal density function for the convolution applied to a learning architecture.

In the experimental part~(Sect.~\ref{sec:EXPRES}), we test different kernel sizes of the weighted convolution, i.e.,~$3 \times 3$,~$5 \times 5$, and~$7 \times 7$. The results of our optimisation model show that the optimal density function reaches better results than the uniform density function, with a reduction of the loss function of the learning model for an average value of~$30\%$. For example, the optimal density function~$\bm{\Phi}$ induced by the generating vector~$\bm{\alpha} = [0.38, 2.21, 1, 2.21, 0.38]$ as~$\bm{\Phi} = \bm{\alpha} \bm{\alpha}^{\top}$ and applied to a~$5\times5$ kernel in the convolution operator reduces the loss function of the learning model of the~$53\%$ with respect to the uniform density function induced by~$\bm{\alpha} = [1, 1, 1, 1, 1]$. We analyse the robustness of the density function to different hyperparameters of the learning model, such as the number and size of the images of the dataset, training epochs, and trainable parameters. We observe that the density function tends to its optimal values when we increase the complexity of the learning model (e.g., number of images, trainable parameters). In contrast, when we reduce the complexity of the network and the data information, the density function tends to a more uniform shape, thus reducing the locality of the convolution. 

We apply our weighted convolution for the solution of a multi-label classification problem and compare our optimal density function with different density functions (e.g., uniform, linear, Gaussian). The optimal density function produces a classification accuracy of~$53\%$, while the uniform density function has a classification accuracy of~$46\%$. We compare the execution time of the convolution with and without the density function, and the computation of the optimal density function. The application of the density function to the convolution increases the execution time by an average of~$11\%$. Finally, we discuss the conclusions and future work~(Sect.~\ref{sec:CONCLUSION}), as the application of the weighted convolution to real-case deep learning problems with 2D and 3D images.

As main contributions, we introduce the weighted convolution and its application to convolutional neural network models. We define an optimisation framework to compute the optimal values of the density function, allowing the model to learn the kernel weights, accounting for the influence of each neighbouring pixel while retaining the same number of trainable parameters. According to the experimental tests, the weighted convolution effectively captures spatial dependencies and improves convergence behaviour and approximation accuracy with respect to standard convolution. The method is robust to the hyperparameters of the learning model and is computationally feasible for real-world deployment, particularly in medical imaging or autonomous systems domains where accuracy gains justify minor efficiency trade-offs. The PyTorch code for the convolution with density function is available at \url{https://github.com/cammarasana123/weightedConvolution}.

\section{Related work}\label{sec:BGRW}
We introduce the convolutional neural networks in terms of architecture and computational cost. As related work, several optimisations are applied to CNNs concerning hyperparameters, activation functions, loss functions, regularisation, and different combinations of the kernels to improve the convolution. 

\paragraph{Convolutional neural network} 
The first application of the convolution in neural networks is introduced in \emph{Cognitron}~\cite{fukushima1975self}, a multilayered neural network. A convolution layer applies multiple convolutions through a set of operators (i.e., a kernel). Each kernel is composed of a set of variables (i.e., weights) that are optimised until the network approximates the desired output properly. The number and dimension of each kernel depend on the architecture and purpose of the network. The bias is an additional parameter that shifts the output to increase the flexibility of the model. Back-propagation is one method for training a network by optimising the kernel weights and minimising the loss function, which represents the accuracy of the network to predict the desired output. A gradient descent optimisation algorithm~\cite{ruder2016overview} or its variants (e.g., Adam and RMSprop~\cite{zou2019sufficient}) are applied to minimise the loss through the back-propagation of the error from the last up to the first layer of the network. In the CNNs, each convolutional layer includes multiple kernels that are trained in parallel to improve the efficiency of the learning. This design allows the model to extract different and hierarchical features from the input data, from early (e.g., edges, textures, and simple patterns) to deeper layers (e.g., complex shapes and objects). Multilayer neural networks with linear layers are equivalent to single-layer networks. Indeed, the convolution layer is usually followed by non-linear layers, such as pooling operators, activation functions (e.g., rectified linear unit, tangent hyperbolic, and logistic sigmoid), and dropout, to increase the modelling capability of the CNNs. The learning model can be tuned with additional hyperparameters such as stride, padding, number of layers and input/output channels.

\paragraph{CNN architecture optimisations}
The \emph{weighted convolutional neural network ensemble}~\cite{frazao2014weighted} combines the output probabilities of convolutional neural networks, where each network has an associated weight that makes networks with better performance have a greater influence on the classification than networks that performed worse. In~\cite{abbood2022new}, a \emph{constrained convolution layer} for the CNN model applies a constrained number of weights in each kernel trained in the phase of learning and excludes the other weights. In the \emph{DropOut}~\cite{wan2013regularization} model, a regularisation of the network is applied to the outputs of a fully connected layer where each element of an output layer is kept with probability~$p$, otherwise being set to~$0$ with probability~$(1-p)$. Regularisation methods~\cite{santos2022avoiding},~\cite{hou2019weighted},~\cite{zheng2018improvement} are designed to reduce over-fitting in CNNs. The optimisation of the hyperparameters usually depends on the type of application, data set, and task. Several works propose an optimisation of the hyperparameters for sensor-based human activity recognition~\cite{raziani2022deep}, diabetic maculopathy diagnosis in optical coherence tomography and fundus retinography~\cite{atteia2022cnn}, and learning-centred emotion recognition for intelligent tutoring systems~\cite{zatarain2020hyperparameter}. Self-attention networks~\cite{vaswani2017attention} learn to focus on different parts of an input sequence when making predictions; instead of processing the sequence in a fixed order, self-attention allows the model to consider the relationships between all elements simultaneously.

\paragraph{Kernel-based convolution and weighted transform}
In~\cite{jia2022variable}, the variables of the kernel are increased with a weighted learnable parameter. The weighting coefficient is not the argument of an optimisation, but rather a parameter given to control the impact of the additional weight on the weight of the kernel. Furthermore, the added parameter is an additional variable for each element of the kernel, thus increasing the computational cost of the training. \emph{Dynamic convolution}~\cite{chen2020dynamic} aggregates multiple parallel convolution kernels dynamically based upon their attentions, which are input dependent. The convolution operator itself maintains a uniform basis function, without any weight. \emph{Omni-dimensional dynamic convolution}~\cite{li2022omni} addresses a multi-dimensional attention mechanism with a parallel strategy to learn complementary attentions for convolutional kernels along all four dimensions of the kernel space at any convolutional layer. In~\cite{ghiasi2019generalizing}, the authors define generalised convolution operators based on positive definite kernel functions to replace the inner product, e.g., non-isotropic Gaussian or Laplacian kernel. In convolutional kernel networks (CKNs)~\cite{mairal2014convolutional}, each layer applies a local kernel approximation to extract features from spatial regions (e.g., patches of an image). Instead of using fixed kernels (e.g. radial basis functions), CKNs learn adaptive kernels from the data. CKNs are extended to graph-based signals~\cite{chen2020convolutional}, where the kernels represent a graph as a feature vector by counting the number of occurrences of some local connected sub-structures.

In~\cite{crandall1994discrete}, the \emph{discrete weighted transforms} are defined as a variant of the fast Fourier transform that includes weighting. \emph{Weighted support vector machine} (WSVM)~\cite{yang2005weighted} assigns different weights to different data points such that the WSVM training algorithm learns the decision surface according to the relative importance of data points in the training data set. The weights in WSVM are generated by the \emph{kernel-based fuzzy c-means} algorithm~\cite{zhang2003kernel}, whose partition generates relatively high values for important data points but low values for outliers. In~\cite{cammarasana2023learning}, the learning of the optimal threshold values for the singular values decomposition is applied to image denoising to reduce the high-frequency components and preserve the main features and contours of the ground-truth image. Weighted wavelet transform~\cite{liu2008weighted} improves lifting and contextually maintains the consistency between the predict and update steps, preserving the reconstruction accuracy. The directional adaptive interpolation~\cite{jiang2002new} improves the orientation property of the interpolated image by adapting to the statistical properties of each image.

\paragraph{Computational aspects} 
Given an input image of size~$N \times N$,~$F$ filters of size~$K\times K$, and~$C$ input channels, the computational cost of the convolution operator applied to the image is~$\mathcal{O}(N^2 \times C \times F \times K^2)$. Common machine learning libraries (e.g., PyTorch~\cite{stevens2020deep}) apply different fast schemes~\cite{lavin2016fast} to reduce the computational cost:
\begin{itemize}
\item \emph{Fast Fourier transform} (FFT) convolution accounts for the convolution theorem, where the convolution in the spatial domain is equivalent to the multiplication in the frequency domain. After the application of the FFT to the image and kernel, the FFT convolution multiplies the frequency representations and then applies the inverse FFT. This approach is usually applied for large kernels with a computational cost of~$\mathcal{O}( N^2 logN)$.
\item \emph{Winograd} algorithm reduces the number of multiplications and increases the number of additions by transforming the convolution into a set of smaller matrix multiplications. This approach is usually applied for small kernels~(e.g.,~$ K = 3$) with a computational cost of~$\mathcal{O}(N^2CFK^2/2)$.
\item \emph{GEMM}-based method unrolls the image into column vectors, reshapes the kernels into a matrix, performs matrix-matrix multiplication (GEMM), and reshapes the result back into the image. While the computational cost is the same as direct convolution, GEMM has optimal performance with BLAS implementation.
\end{itemize}
%
\section{Weighted convolution and optimised density functions}\label{sec:METHOD}
After the introduction of the standard convolution operator~(Sect.~\ref{sec:WC}), we define the weighted convolution with density function~(Sect.~\ref{sec:DFC}) the optimisation model for the computation of the optimal density function~(Sect.~\ref{sec:WO}), and the related computational cost~(Sect.~\ref{sec:CC}).
\subsection{Convolution operator and learning model}\label{sec:WC}
Given a compact domain~$\Omega \in \mathbb{R}^n$ and two functions~$f, g \in \mathcal{L}^1(\Omega)$, the convolution~$f \ast g$ is defined as
\begin{equation}\label{eq:convCont}
 (f \ast g) (\bm{t}) := \int_{\Omega} f(\bm{\tau}) g(\bm{t}-\bm{\tau}) d\bm{\tau}.
 \end{equation}
In CNNs,~$\Omega$ is a discrete 2D domain and the functions~$f,g$ are the input signal (e.g., the 2D image) and the filter (e.g., the convolution kernel), respectively. Given the input signal~$\bm{I} \in \mathbb{R}^{R \times C}$ defined on a 2D regular grid and the tensor of kernels~$\bm{W} \in \mathbb{R}^{K_a \times K_b \times F}$ composed of F kernels~$\bm{w}$ of size~$K_a \times K_b$ as~$(\bm{W})_{ijf} := \bm{w}_{ij}^f~$, we discretise the convolution in Eq.~(\ref{eq:convCont}) of~$\bm{I}$ with the kernels~$\bm{W}$ as
\begin{equation}\label{eq:CONVOLUTION}
\left\{
\begin{array}{l}
(\bm{I} \ast \bm{W})_{ij}^f := \sum_{a=1}^{K_a} \sum_{b=1}^{K_b} \bm{w}_{ab}^{f} \cdot \bm{I}_{i+a-K_a+1,j+b-K_b+1},\\
i = 1 \ldots R, j = 1\ldots C, f = 1\ldots F.
\end{array}
\right.
\end{equation}
%
%
Introducing the neighbourhood~$\mathcal{N}(\bm{I}_{ij})$ as the~$K_a \times K_b$ sub-matrix of~$\bm{I}$ centred in~$(i,j)$, the discrete convolution is rewritten in matrix form as \mbox{$(\bm{I} \ast \bm{W})_{ij}^f := \langle \bm{w}^f , \mathcal{N}(\bm{I}_{ij}) \rangle_F$}, where
the Frobenius inner product~$\langle \cdot, \cdot \rangle_F$ of two matrices~$\mathbf{A}, \mathbf{B} \in \mathbb{R}^{m \times n}$ is defined as~$\langle \mathbf{A}, \mathbf{B} \rangle_F := \sum_{i=1}^{m} \sum_{j=1}^{n} A(i,j) B(i,j)$. 

Given an input~$\mathcal{I} = \left\{\bm{I}_i\right\}_{i=1}^N$ and target~$\mathcal{T} = \left\{\bm{T}_i\right\}_{i=1}^N$ data set of~$N$ images, we define a learning model~$\mathcal{M}$ as the minimisation of a loss function~$\mathcal{L}$ with respect to the kernels~$\bm{W}$ (i.e., the weights) between the output data set and the output of the network~$\hat{\mathcal{T}}$
\begin{equation}\label{eq:learningModel}
\mathcal{M}:\qquad \min_{\bm{W}} \mathcal{L}( \mathcal{T} , \hat{\mathcal{T}}(\bm{W}) ),
\end{equation}
where~$\hat{\mathcal{T}}$ is the output of the combination of convolution layers in Eq.~(\ref{eq:CONVOLUTION}) and non-linear operators applied to the input~$\mathcal{I}$.

\subsection{Weighted convolution and learning model\label{sec:DFC}}
Given the density function~$\varphi \in \mathcal{L}^{\infty}(\Omega)$, we introduce the \emph{weighted convolution} as
\begin{equation*}
(f \ast g_\varphi)(\bm{t}) := \int_{\Omega}  f(\bm{\tau}) (\varphi(\bm{t}-\bm{\tau})  g(\bm{t}-\bm{\tau})) d\bm{\tau}.
\end{equation*}
The weighted convolution reduces to the standard convolution when \mbox{$\varphi:=1$}. Discretising the density function~$\varphi$ on a 2D regular grid as~$\bm{\Phi} \in \mathbb{R}^{K_a \times K_b}$, we define the tensor of kernels with density function as~$\bm{W}_{\bm{\Phi}} \in \mathbb{R}^{K_a \times K_b \times F}$, where~$(\bm{W}_{\bm{\Phi}})_{ijf} := \bm{\Phi}_{ij} \bm{w}_{ij}^f~$, and the \emph{discrete weighted convolution} as
\begin{equation}\label{eq:wCONVOLUTION}
\left\{
\begin{array}{l}
(\bm{I} \ast \bm{W}_{\bm{\Phi}})_{ij}^{f} := \sum\limits_{a=1}^{K_a} \sum\limits_{b=1}^{K_b} (\bm{\Phi}_{ab} \bm{w}_{ab}^{f} ) \bm{I}_{i+a-K_a+1,j+b-K_b+1},\\
i = 1 \ldots R,\, j = 1\ldots C,\, f = 1\ldots F.
\end{array}
\right.
\end{equation}
In matrix form, \mbox{$(\bm{I} \ast \bm{W}_{\bm{\Phi}})_{ij}^{f} := \langle \bm{\Phi} \circ \bm{w}^f, \mathcal{N}(\bm{I}_{ij} )\rangle_F$}, where the Hadamard product~$\circ$ is the component-wise product of two matrices, i.e.,~$\langle\mathbf{A},\mathbf{B}\rangle_{F}=:\mathbf{C}$,~$C(i,j):=A(i,j)B(i,j)$,~$\mathbf{A},\,\mathbf{B},\,\mathbf{C}\in\mathbb{R}^{m\times n}$. The density function~$\bm{\Phi}$ is shared across both the image and the kernels. We underline that the discrete weighted convolution with a uniform density function~$\bm{\Phi} = \bm{1}$ is equivalent to the discrete convolution without a density function. Table~\ref{tab:convComparison} compares standard and weighted convolution.
\begin{table}[t]
\caption{Comparison between standard and weighted convolution.\label{tab:convComparison}}
\begin{tabular}{c}
\includegraphics[width=1.0\textwidth]{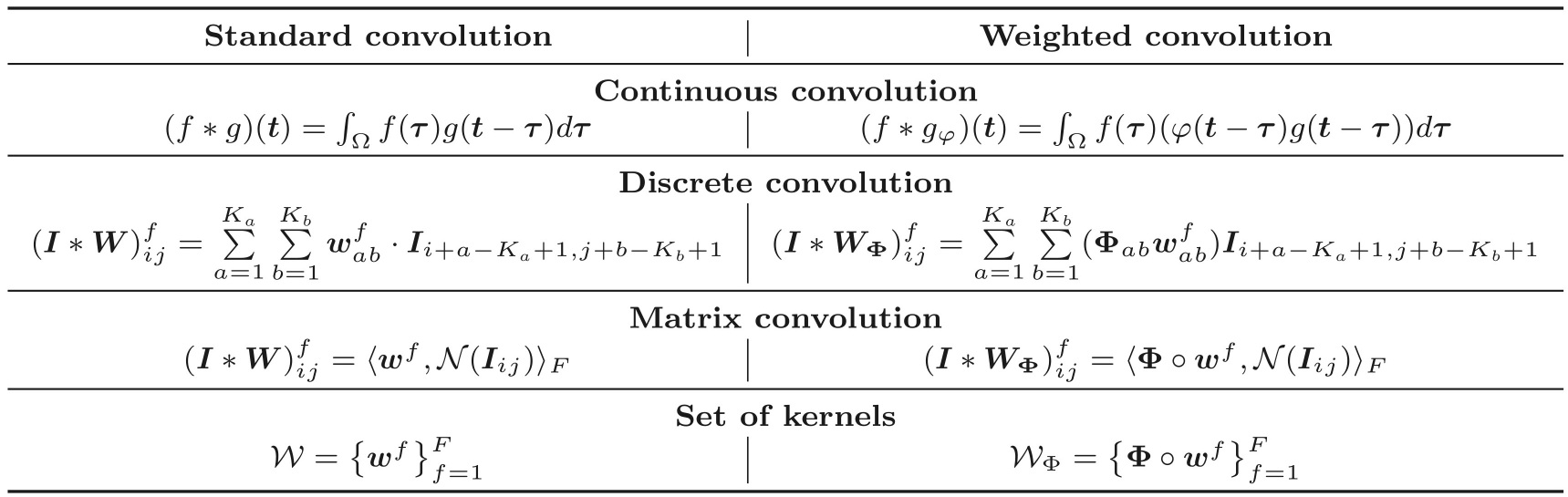}
\end{tabular}
\end{table}
Given the learning model in Eq.~(\ref{eq:learningModel}) and replacing the standard convolution~$\bm{I} \ast \bm{W}$ in Eq.~(\ref{eq:CONVOLUTION}) with the convolution with the density function~$\bm{I} \ast \bm{W}_{\bm{\Phi}}$ in Eq.~(\ref{eq:wCONVOLUTION}), we define the learning model
\begin{equation}\label{eq:wModel}
\mathcal{M}_{\bm{\Phi}}:\qquad\min_{\bm{W}} \mathcal{L}( \mathcal{T} , \hat{\mathcal{T}}(\bm{W}_{\bm{\Phi}}) ).
\end{equation}
\subsection{Density function optimisation}\label{sec:WO}
To compute the \emph{optimal density function} (i.e., the values of~$\bm{\Phi}$) of the learning model in Eq.~(\ref{eq:wModel}), we introduce the optimisation model as
\begin{equation}\label{eq:FUNCTIONAL}
\min_{\bm{\Phi}} \mathcal{M}_{\bm{\Phi}}, 
\end{equation}
where we minimise the objective function through the values of the density function~$\bm{\Phi}$, and the objective function~$\mathcal{M}_{\bm{\Phi}}$ is computed through the solution of Eq.~(\ref{eq:wModel}), where we minimise the loss function of the learning model. Given a squared kernel~$K_a = K_b = K$,~$K$ odd,~$m = (K + 1)/2$, we define the discrete density function as the~$K_a \times K_b$ matrix~$\bm{\Phi} = \bm{\alpha} \bm{\beta}^{\top}$ (i.e.,~$ \bm{\Phi}_{ij} = \bm{\alpha}_i \bm{\beta}_j$), with~$\bm{\alpha} \in \mathbb{R}^{K_a}, \bm{\beta} \in \mathbb{R}^{K_b}$, where~$\bm{\alpha}$ and~$\bm{\beta}$ represent the scaling factors along the two dimensions of the regular grid. We assume the following properties for the density function:
\begin{itemize}
\item Symmetry along both the dimensions, i.e.,~$\bm{\alpha} = \bm{\beta}$,~$\bm{\alpha}(i) = \bm{\alpha}(K - i + 1), i = 1\ldots K$;
\item Value of the central node as~$\bm{\alpha}_m = M$.
\end{itemize}
The density function~$\bm{\Phi} \in \mathbb{R}^{K \times K}$ is defined through~$(K-1)/2$ coefficients, is symmetric, positive semidefinite, and with rank~$1$. Indeed, we reduce the computation of the density function to the coefficients of~$\bm{\alpha}$, as~$\bm{\Phi} = \bm{\alpha} \bm{\alpha}^{\top}$.
\begin{figure*}[t]
\centering
\begin{tabular}{c|cc}
\includegraphics[width=.3\textwidth]{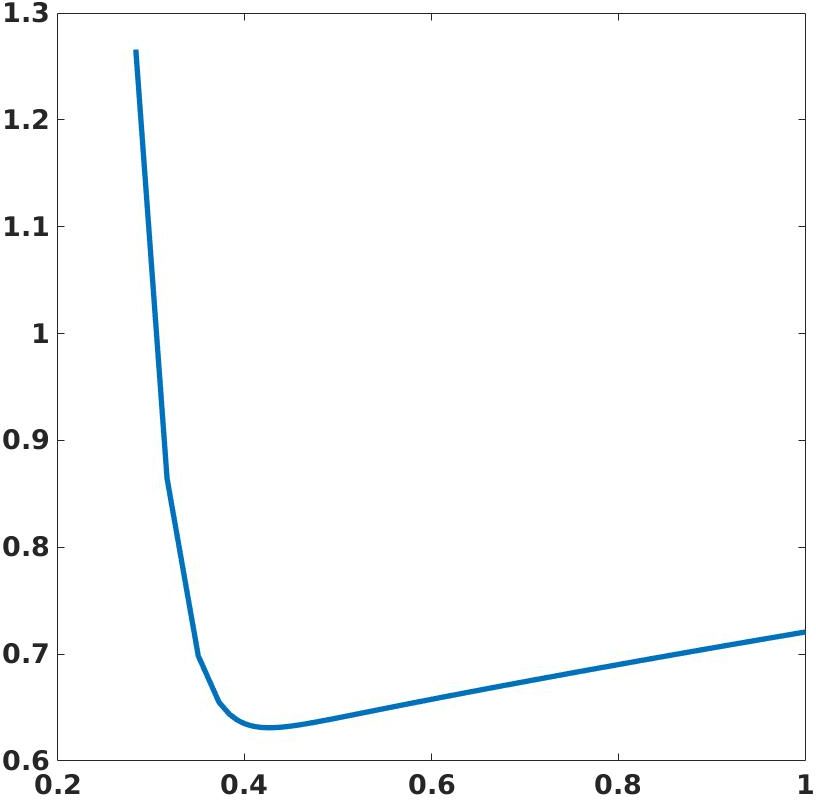} &
\includegraphics[width=.3\textwidth]{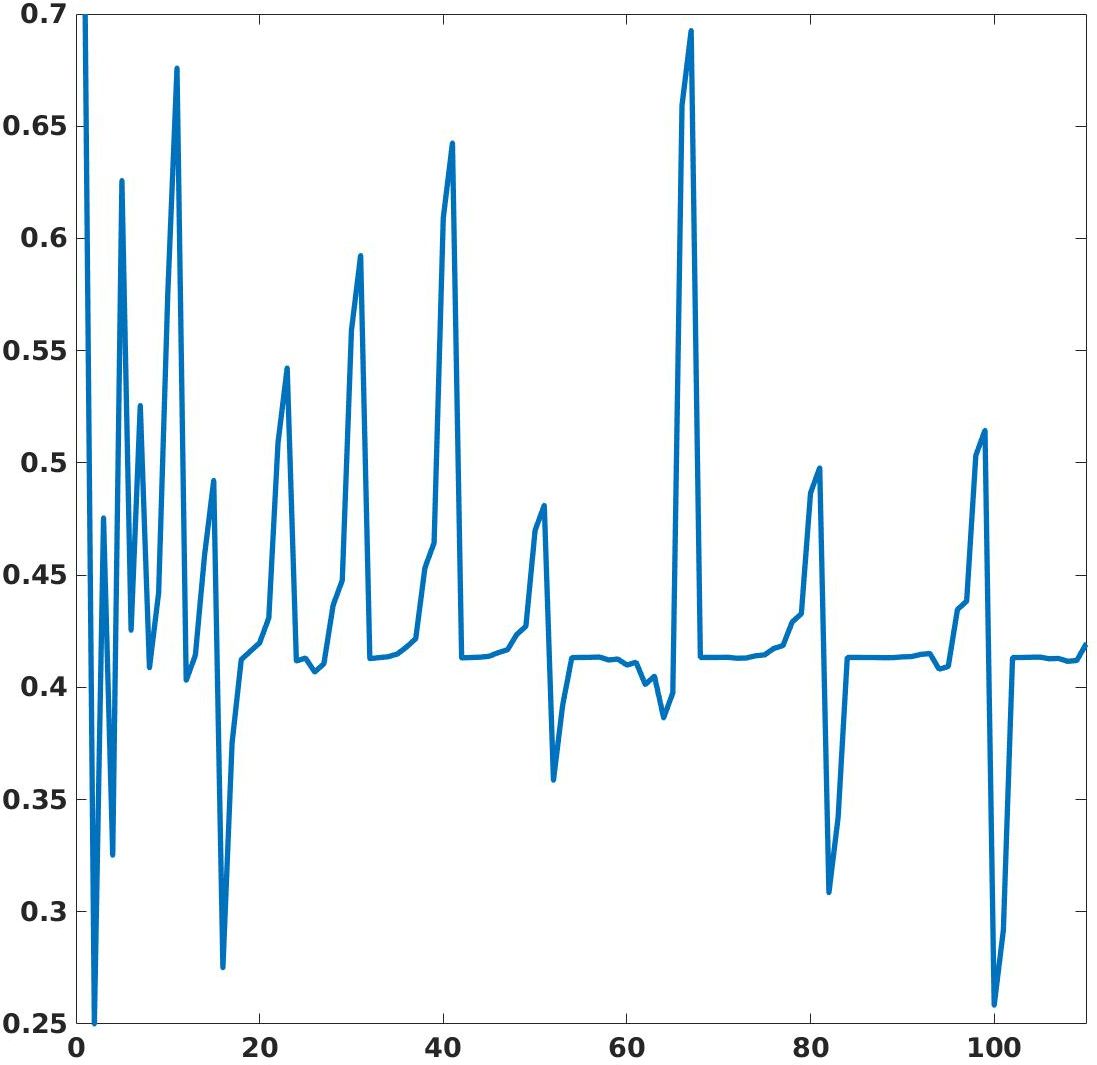} &
\includegraphics[width=.3\textwidth]{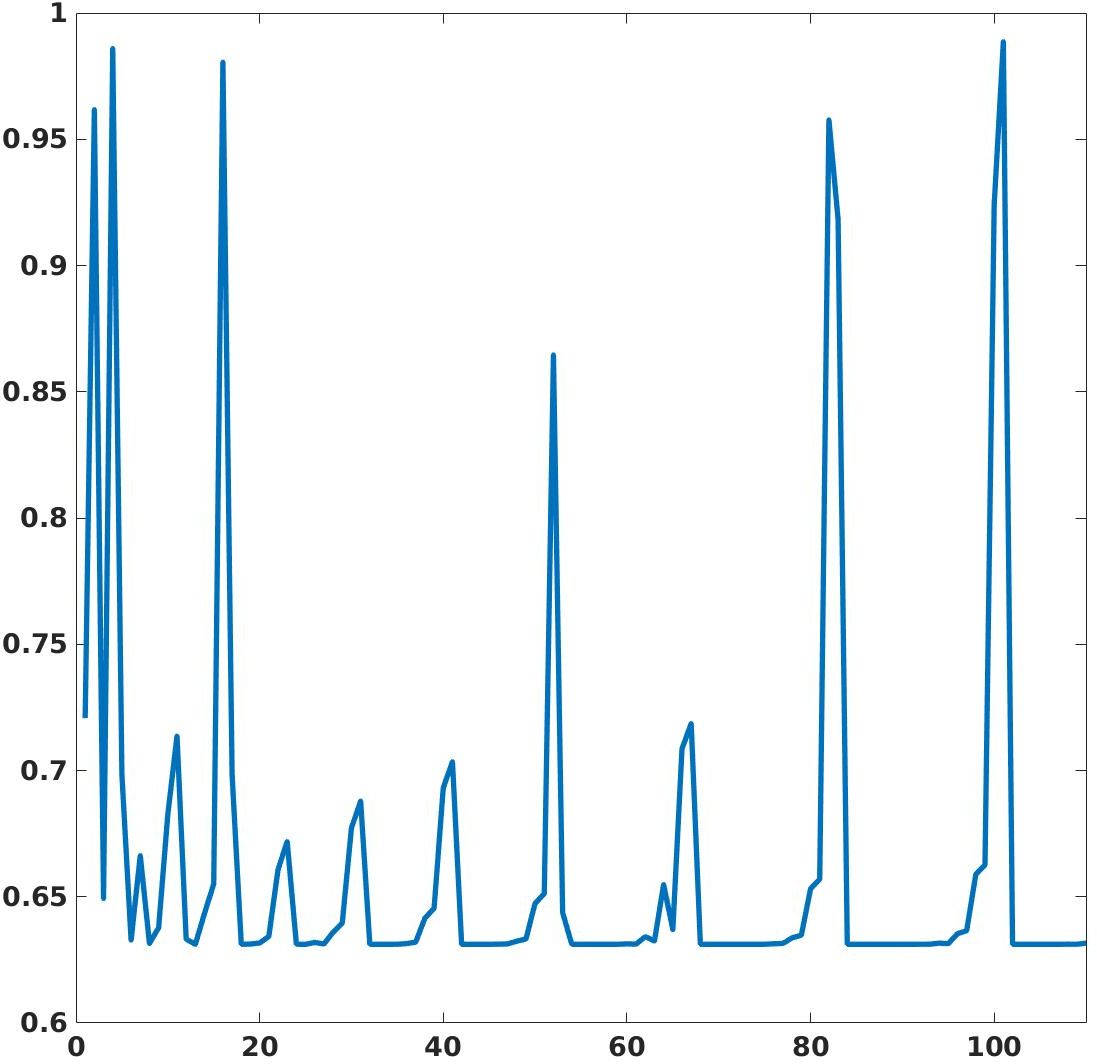} 
\end{tabular}
\caption{Optimisation results on the~$3\times3$ kernel. Left:~$\bm{\alpha}_1$ values ($x-$axis) with respect to the objective function ($y-$axis). Centre:~$\bm{\alpha}_1$ optimisation ($y-$axis) over the iterations of the optimiser ($x-$axis). Right: objective function ($y-$axis) over the iterations of the optimiser ($x-$axis).\label{FIG:3x3}}
\end{figure*}
\paragraph{Learning model set-up and optimisation algorithms}
We define an \emph{image-to-image learning model}, where the input is a noisy image and the output is the corresponding denoised image. The learning model is composed of three convolution layers, with:~$L_{1,c}^{1}$,~$L_{c,c}^{2}$,~$L_{c,1}^{3}$, where given~$L_{i,o}^{l}$ we assign several input~$i$ and output~$o$ channels for each layer~$L$. We set the stride of~$L^1$ to~$s$,~$L^2$ to~$1$, and~$L^3$ to~$1/s$, which corresponds to a transposed convolution layer of stride~$s$. We perform batch normalisation and a rectified unit activation after each convolutional layer. The loss function is computed as the mean squared error between the predicted and the ground-truth image. 

Our optimisation model in Eq.~(\ref{eq:FUNCTIONAL}) separates the optimisers applied for the computation of the kernel weights (i.e.,~$\bm{W}$) and the density function values (i.e.,~$\bm{\alpha}$). We select DIRECT-L as the method for the computation of the optimal density function. DIRECT~\cite{jones1993lipschitzian} is a global, derivative-free, and deterministic search algorithm that systematically divides the search domain into smaller hyper-rectangles. Rescaling the bound constraints to a hypercube gives equal weight to all dimensions in the search procedure. DIRECT derives from Lipschitzian global optimisation, i.e., a branch-and-bound model, where bounds are computed through the knowledge of a Lipschitz constant for the objective function. DIRECT introduces modifications to the Lipschitzian approach to improve the results in higher dimensions by eliminating the need to know the Lipschitz constant. The global optimiser DIRECT-L~\cite{gablonsky2000locally} is the locally-biased form that enhances the efficiency of functions without too many local minima. DIRECT-L supports linearly-constrained problems, does not require analytic or numeric derivatives to compute the optimal solution, and applies a global search of the optimal solution, which is required as the functional may lack of strictly convexity property~\cite{cammarasana2025analysis}. In our framework, we define~$[0,2M]$ as lower and upper bounds for the values of~$\bm{\alpha}$, and an initialisation value of~$M$ for each element of~$\bm{\alpha}$. 

We select SGD~\cite{amari1993backpropagation} as the method for the minimisation of the loss function in the learning model and the computation of the optimal weights of the kernel within the convolution. The SGD is a local method that applies to differentiable functions by updating parameters in the direction of the negative gradient. In our framework, the weights are initialised with the Kaiming initialisation~\cite{he2015delving}. 

\subsection{Computational cost}\label{sec:CC}
Recalling that~$K_a = K_b = K$, the computational cost of Eq.~(\ref{eq:CONVOLUTION}) is~$\mathcal{O}(RCF(2K^2))$. Instead, the computational cost of Eq.~(\ref{eq:wCONVOLUTION}) is~$\mathcal{O}(RCF(3K^2))$. In fact, for each kernel element, the convolution with the density function adds one multiplication with respect to the convolution without the density function. The optimal density function is computed by minimising the functional in Eq.~(\ref{eq:FUNCTIONAL}) with the proposed learning model and the constraints on the density function properties. The minimisation of the functional includes the solution of the learning model. We underline that the kernel weights are the variables of the learning model, while the values of the density function at its nodes are the variables of the functional minimisation. With our definition of the density function properties, the optimisation model in Eq.~(\ref{eq:FUNCTIONAL}) with a~$3 \times 3$ kernel size requires only one variable, as~$\bm{\alpha} \in \mathbb{R}^3$,~$\bm{\beta} = \bm{\alpha}$,~$\bm{\alpha}_2 = M$, and~$\bm{\alpha}_1 = \bm{\alpha}_3$. Generally, it requires~$(K-1)/2$ variables. Given~$\bm{\alpha} \in \mathbb{R}^{n}$, the DIRECT-L method requires, as the worst case, a computational cost of~$\mathcal{O}(2^n)$, even if novel variants of this method propose improved performances.

\section{Experimental results}\label{sec:EXPRES}
We discuss the computation of the optimal density function (Sect.~\ref{sec:EXPRES1}) and the execution time (Sect.~\ref{sec:EXPRES2}). As an abuse of definition, we refer to~$\bm{\alpha}$ as the density function, in the definition of~$\bm{\Phi} = \bm{\alpha} \bm{\alpha}^{\top}$.
\begin{figure*}[t]
\centering
\begin{tabular}{c|cc}
\includegraphics[width=.3\textwidth]{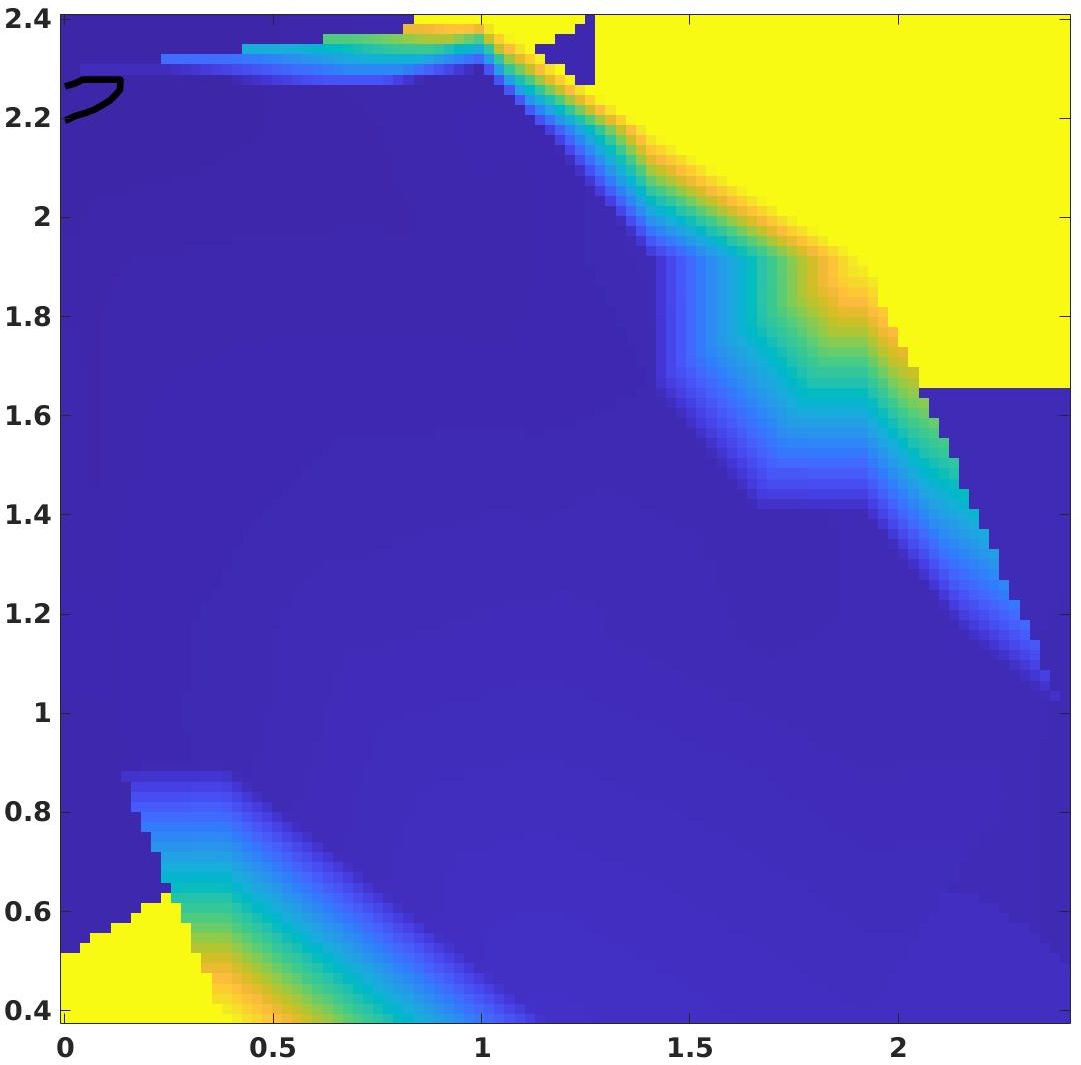} &
\includegraphics[width=.3\textwidth]{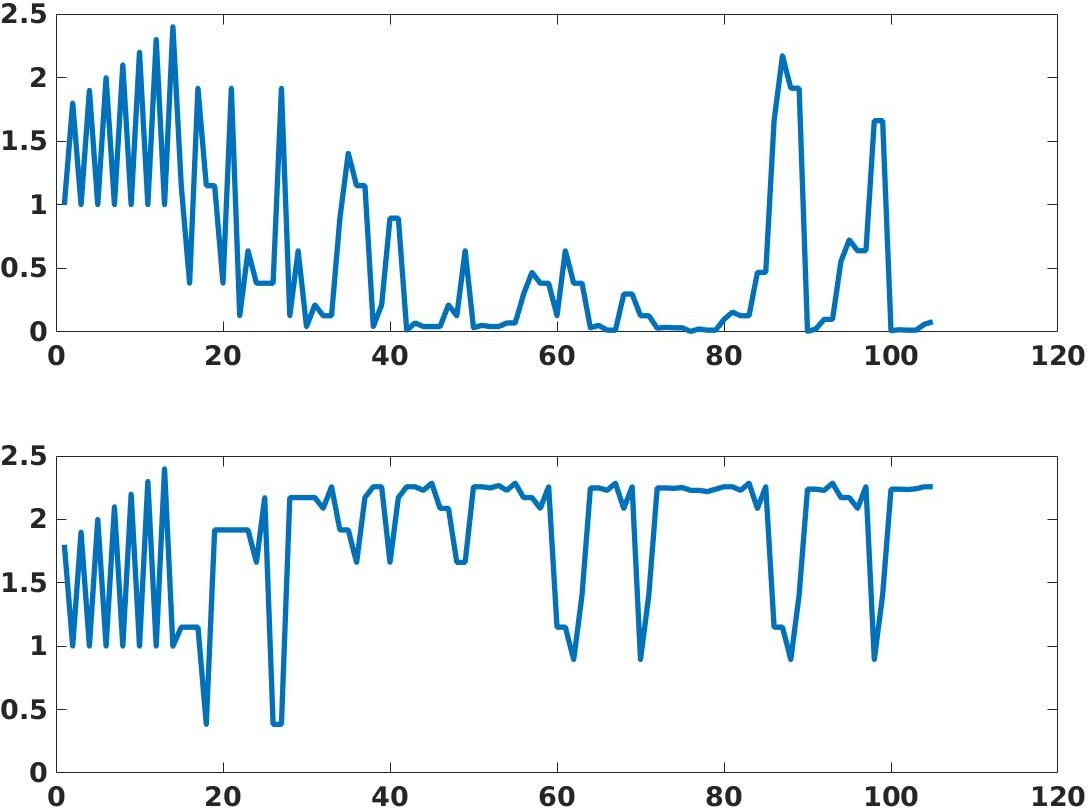} &
\includegraphics[width=.3\textwidth]{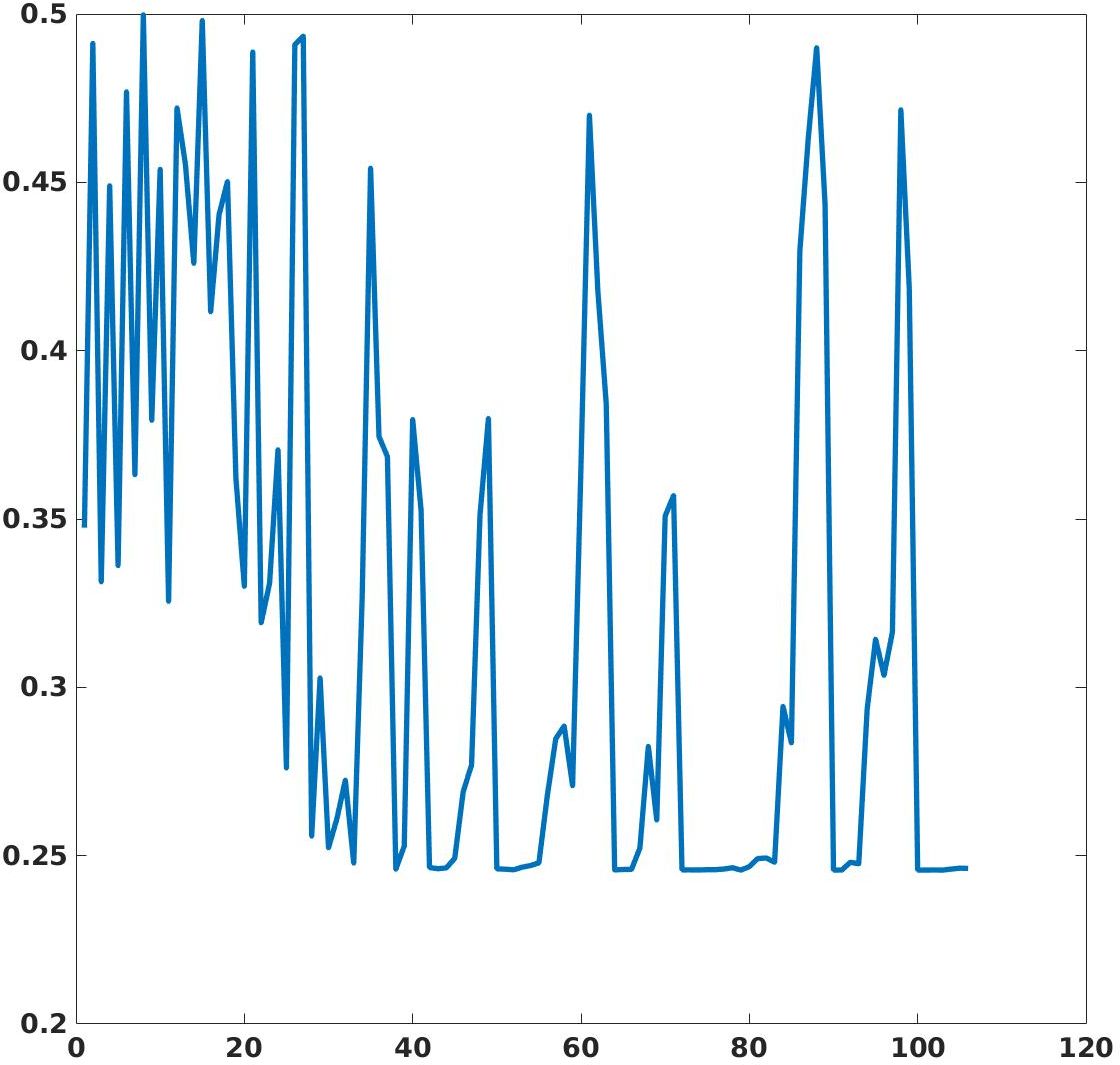} 
\end{tabular}
\caption{Optimisation results on the~$5\times5$ kernel. Left:~$\bm{\alpha}_1$ ($x-$axis) and~$\bm{\alpha}_2$ ($y-$axis) values with respect to the objective function: from blue (low) to yellow (high), and black iso-contour of minimal values of the loss function. Centre:~$\bm{\alpha}$ values optimisation ($y-$axis) over the iterations of the optimiser ($x-$axis): top graph represents~$\bm{\alpha}_1$, bottom graph represents~$\bm{\alpha}_2$. Right: objective function ($y-$axis) over the iterations of the optimiser ($x-$axis).\label{FIG:5x5}}
\end{figure*}
\subsection{Optimal density function}\label{sec:EXPRES1}
We test a~$3 \times 3$,~$5 \times 5$, and~$7 \times 7$ kernel size. The~$3 \times 3$ kernel size is commonly used in modern CNNs (e.g., VGG~\cite{simonyan2014very}), while a larger kernel size is often used in initial layers to capture broader spatial context (e.g., the first layer of ResNet-50~\cite{koonce2021resnet} is designed with a~$7 \times 7$ kernel). We solve the proposed learning model~$\mathcal{M}_{\bm{\Phi}}$ in Eq.~(\ref{eq:FUNCTIONAL}) for 20 epochs, with an SGD optimiser with a learning rate of~$0.01$. The data set is composed of~$200$ images with a~$256 \times 256$ resolution. We set the stride~$s = 1$, and the number of channels~$c = 4$.
\begin{figure*}[t]
\centering
\begin{tabular}{c|cc}
\includegraphics[width=.3\textwidth]{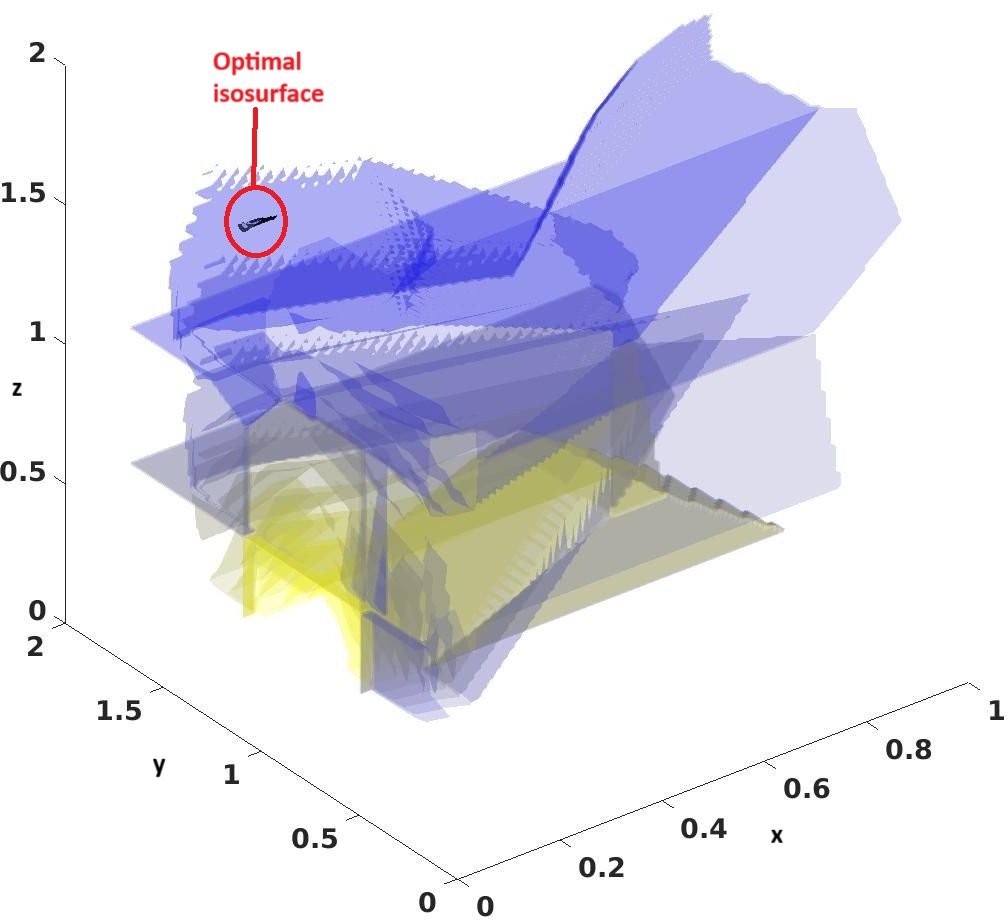} &
\includegraphics[width=.3\textwidth]{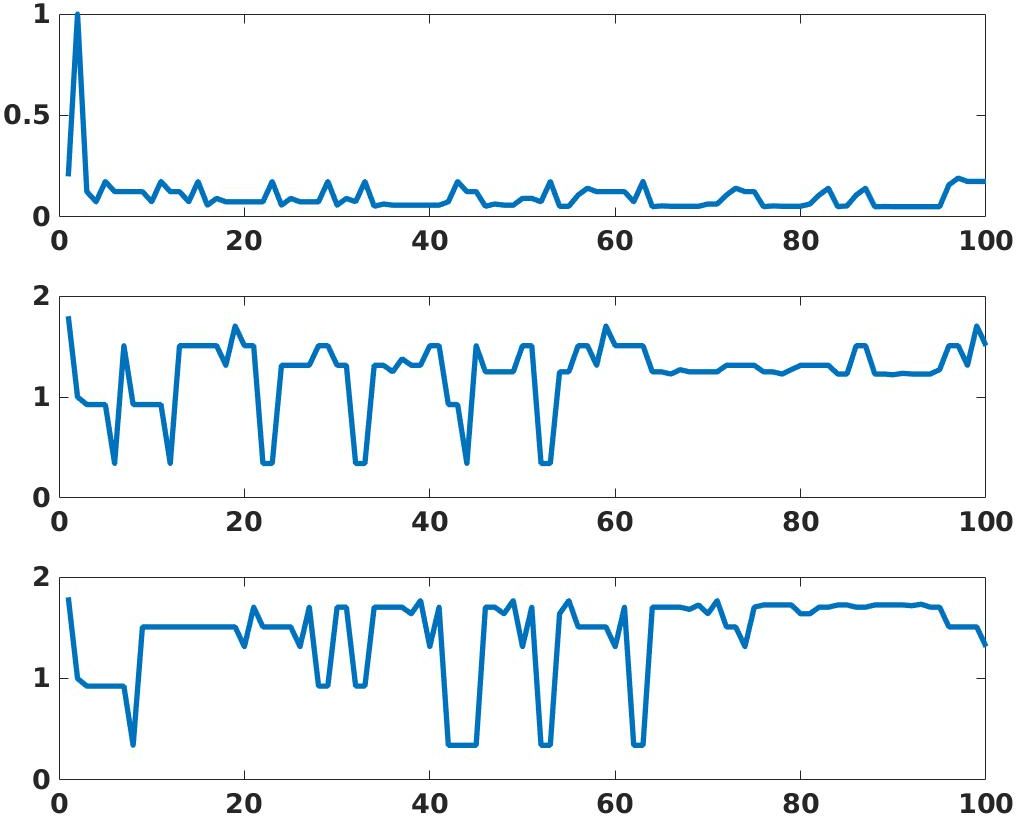} &
\includegraphics[width=.3\textwidth]{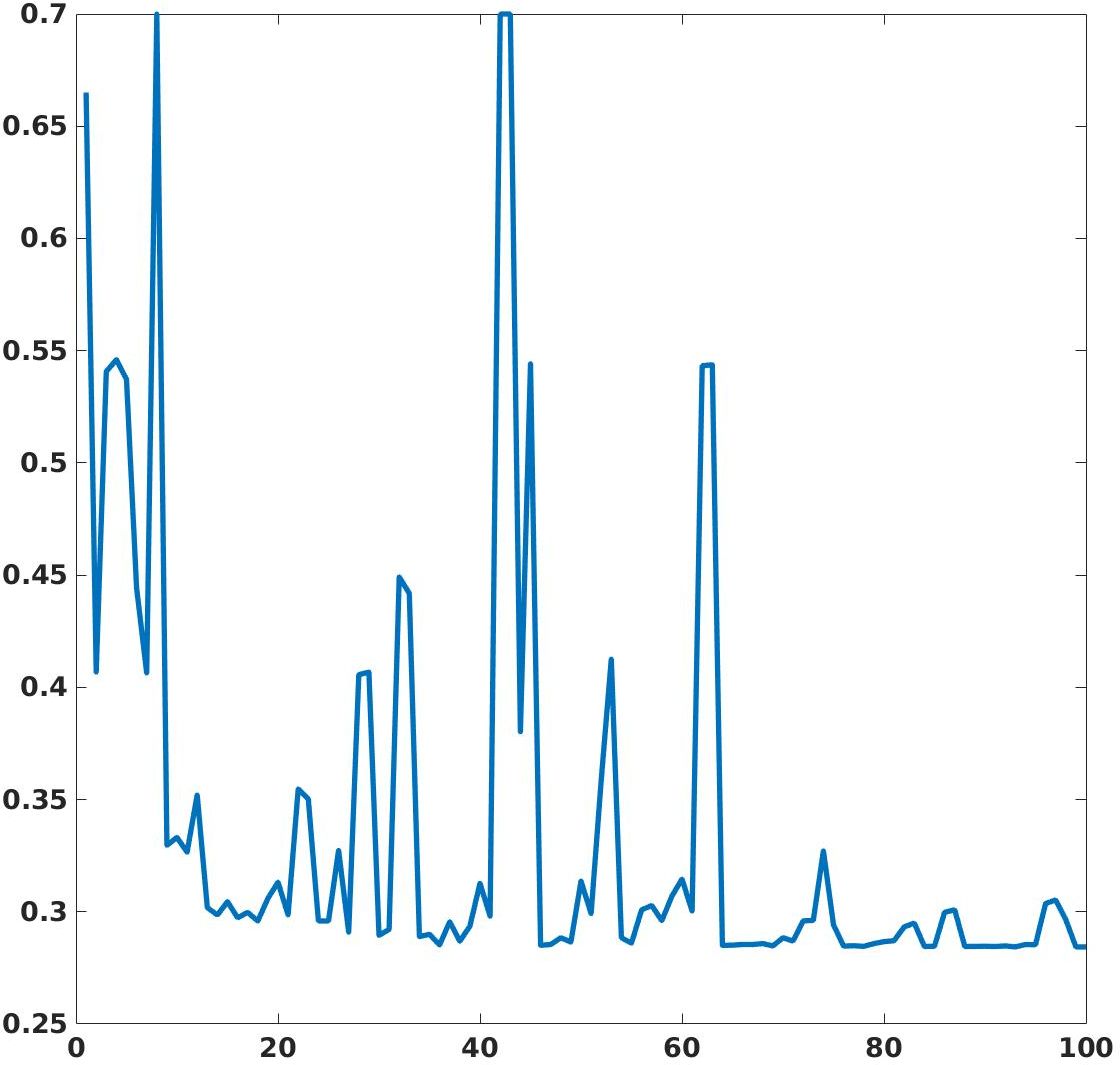} 
\end{tabular}
\caption{Optimisation results on the~$7\times7$ kernel. Left:~$\bm{\alpha}_1$ ($x-$axis),~$\bm{\alpha}_2$ ($y-$axis), and~$\bm{\alpha}_3$ ($z-$axis) values with respect to the objective function: from blue (low) to yellow (high), and black iso-surface of minimal values of the loss function. Centre:~$\bm{\alpha}$ values optimisation ($y-$axis) over the iterations of the optimiser ($x-$axis): top graph represents~$\bm{\alpha}_1$, middle graph represents~$\bm{\alpha}_2$, and bottom graph represents~$\bm{\alpha}_3$. Right: objective function ($y-$axis) over the iterations of the optimiser ($x-$axis).\label{FIG:7x7}}
\end{figure*}

The values of the density function are initialised with~$M=1$ and the absolute tolerance value of the functional minimum of DIRECT-L to~$1 \times 10^{-6}$. We compute the optimal values of the density function in the form~$\bm{\alpha} = [\bm{\alpha}_1, 1, \bm{\alpha}_1]$ for the~$3 \times 3$ kernel,~$\bm{\alpha} = [\bm{\alpha}_1,\bm{\alpha}_2, 1, \bm{\alpha}_2, \bm{\alpha}_1]$ for the~$5 \times 5$ kernel, and~$\bm{\alpha} = [\bm{\alpha}_1, \bm{\alpha}_2, \bm{\alpha}_3, 1, \bm{\alpha}_3, \bm{\alpha}_2, \bm{\alpha}_1]$ for the~$7 \times 7$ kernel. Figs.~\ref{FIG:3x3},~\ref{FIG:5x5}, and~\ref{FIG:7x7} show the optimisation results for the~$3\times3$,~$5\times5$, and~$7\times7$ kernels respectively. In Fig.~\ref{FIG:3x3}, the optimiser converges to an optimal value of~$\bm{\alpha}_1 \approx 0.42$. The optimal values of the density function~$[0.42, 1, 0.42]$ reduce the objective function of the~$12\%$ with respect to the uniform density function~$[1, 1, 1]$. Fig.~\ref{FIG:3x3}(left) shows that the~$\bm{\alpha}_1$ value has a convex behaviour with respect to the objective function, with the minimum value in~$0.42$. In Fig.~\ref{FIG:5x5}, the optimiser converges to the optimal values of~$\bm{\alpha}_1 \approx 0.38$,~$\bm{\alpha}_2 \approx 2.21$. The optimal density function~$[0.38, 2.21, 1, 2.21, 0.38]$ reduces the objective function of the~$53\%$ with respect to the uniform density function~$[1, 1, 1, 1, 1]$. In Fig.~\ref{FIG:7x7}, the optimiser converges to the optimal values of~$\bm{\alpha}_1 \approx 0.06$,~$\bm{\alpha}_2 \approx 1.23$, and~$\bm{\alpha}_3 \approx 1.72$. The optimal density function~$[0.06, 1.23, 1.72, 1, 1.72, 1.23, 0.06]$ reduces the objective function of the~$30\%$ with respect to the uniform density function~$[1, 1, 1, 1, 1, 1, 1]$.
\begin{figure*}[t]
\centering
\begin{tabular}{ccc}
\includegraphics[width=.3\textwidth]{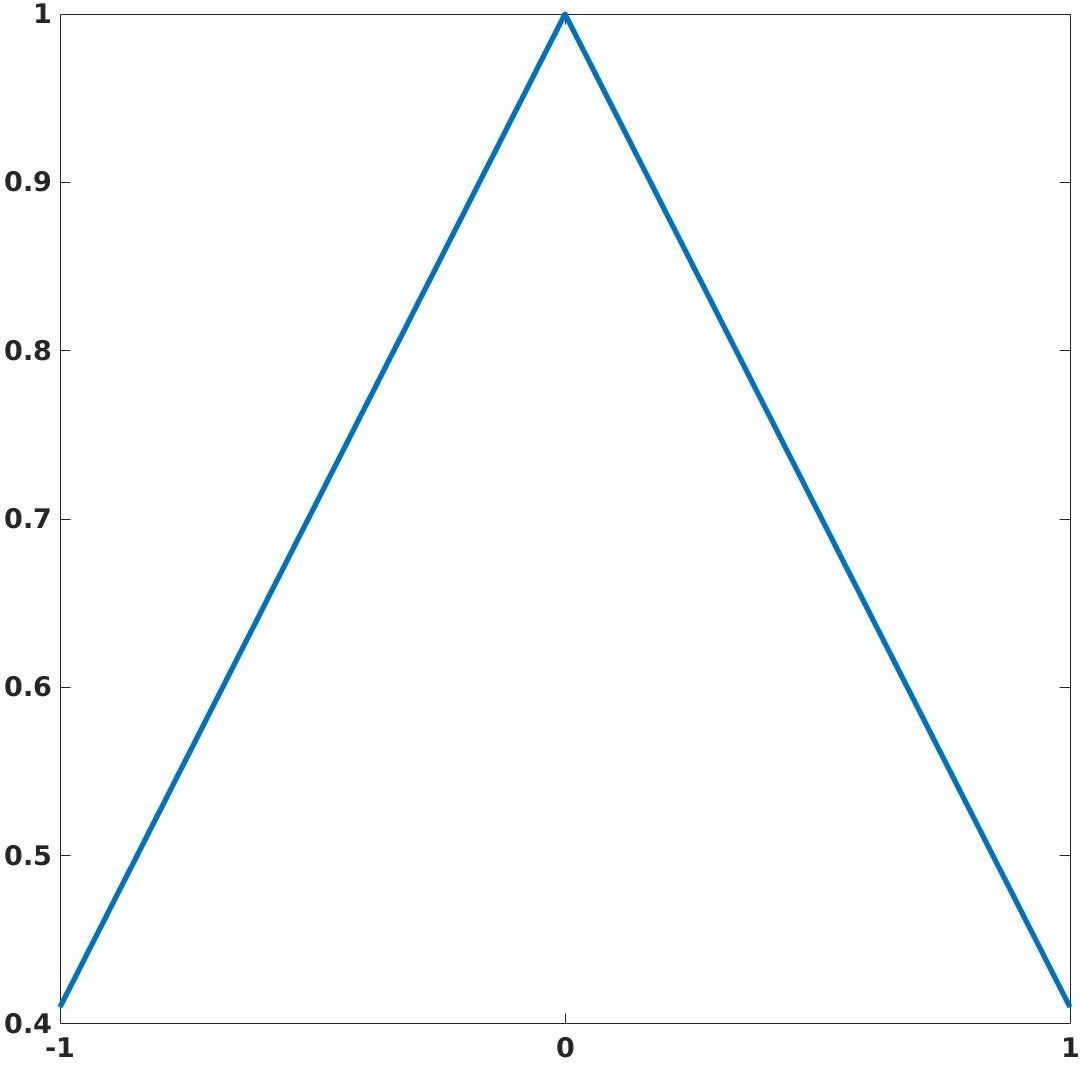} &
\includegraphics[width=.3\textwidth]{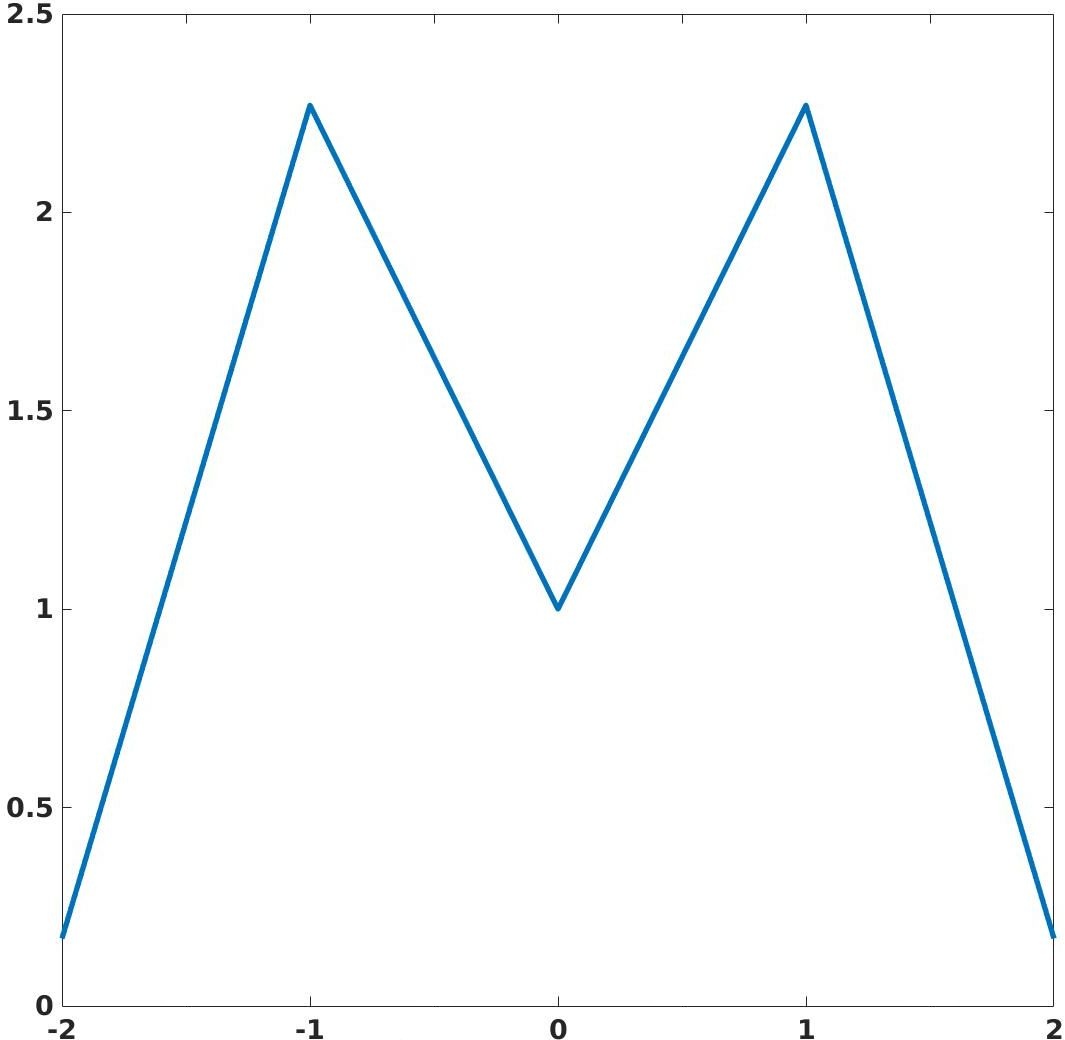} &
\includegraphics[width=.3\textwidth]{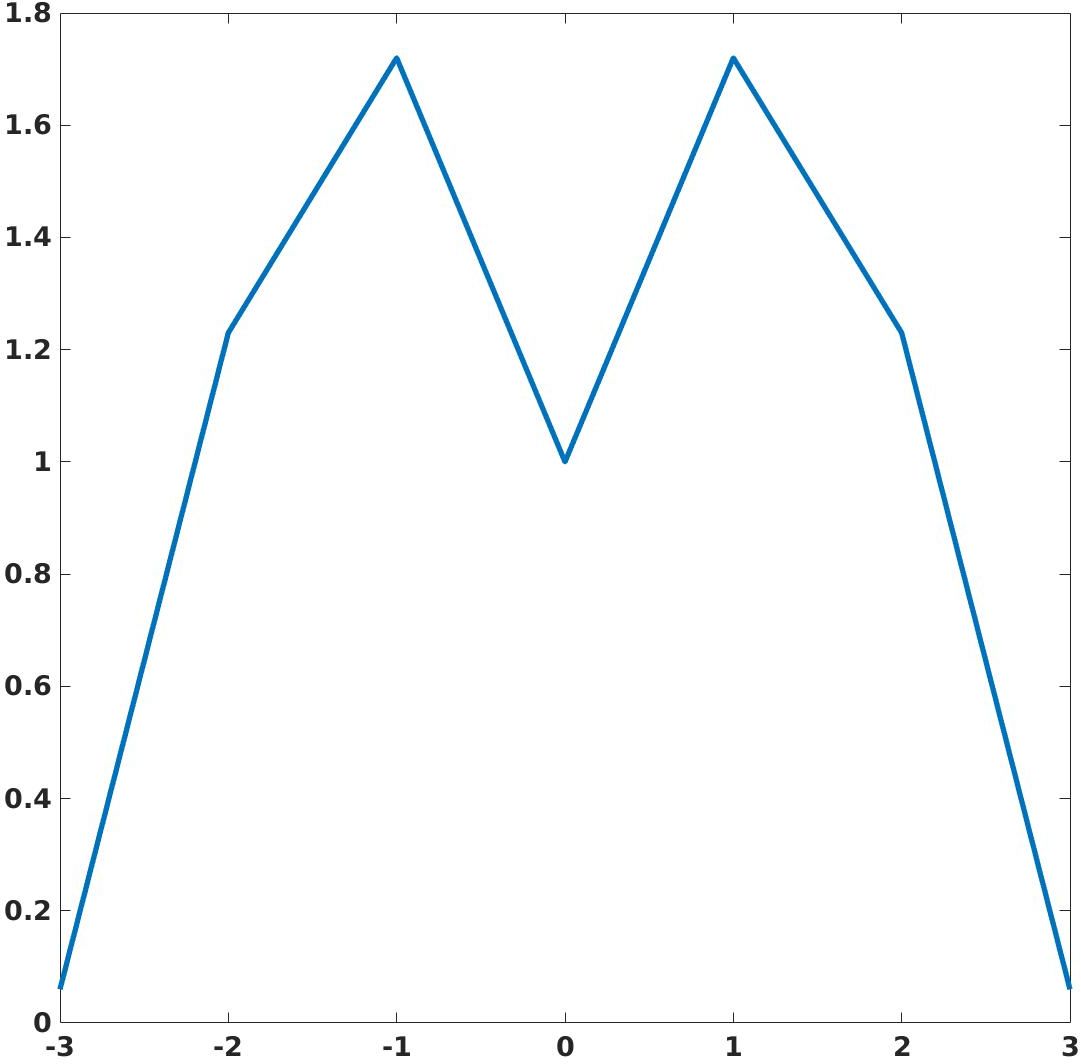} 
\end{tabular}
\caption{$3\times3$ (left),~$5\times5$ (centre), and~$7\times7$ (right) density function. Nodes on the~$x-$axis, density function values on the~$y-$axis.\label{FIG:kernel}}
\end{figure*}

Figs.~\ref{FIG:3x3},~\ref{FIG:5x5}, and~\ref{FIG:7x7} (right side) show the objective function of the optimiser when solving Eq.~(\ref{eq:FUNCTIONAL}). For the clarity of visualisation, we show only the first~$100$ iterations, while the total number of iterations is discussed in Sect.~\ref{sec:EXPRES2}. The peaks of the objective function are generated by the optimiser \emph{DIRECT-L}, which is a global optimiser that does not account for the derivatives of the functional. Therefore, it spans over the full range of values of each variable, generating an irregular behaviour in terms of convergence to the optimal value during the first iterations.

Fig.~\ref{FIG:kernel} shows the optimal density function associated with the~$3\times3$ (left),~$5\times5$ (centre), and~$7\times7$ (right) kernel. We underline the behaviour of the density function in terms of convolution operation: the optimal~$\bm{\alpha} \in \mathbb{R}^3$ density function has a larger value on the central node, while the value on the external nodes are lower; the optimal~$\bm{\alpha} \in \mathbb{R}^5$ density function has a larger value on the nodes adjacent to the central one, and the values on the external nodes are lower; the optimal~$\bm{\alpha} \in \mathbb{R}^7$ density function has a larger value on the nodes adjacent to the central one, has lower values on the~$-2$ and~$+2$ nodes but still above the value of the central node, and then significantly lower values on the external nodes. This shape of the~$\bm{\alpha} \in \mathbb{R}^7$ density function recalls some well-known basis functions with proper parametrisations, such as Catmull-Rom spline~\cite{twigg2003catmull} and Ricker Mexican hat wavelet~\cite{ryan1994ricker}.

\begin{figure}[t]
\centering
\begin{tabular}{cc}
\includegraphics[width=0.40\textwidth]{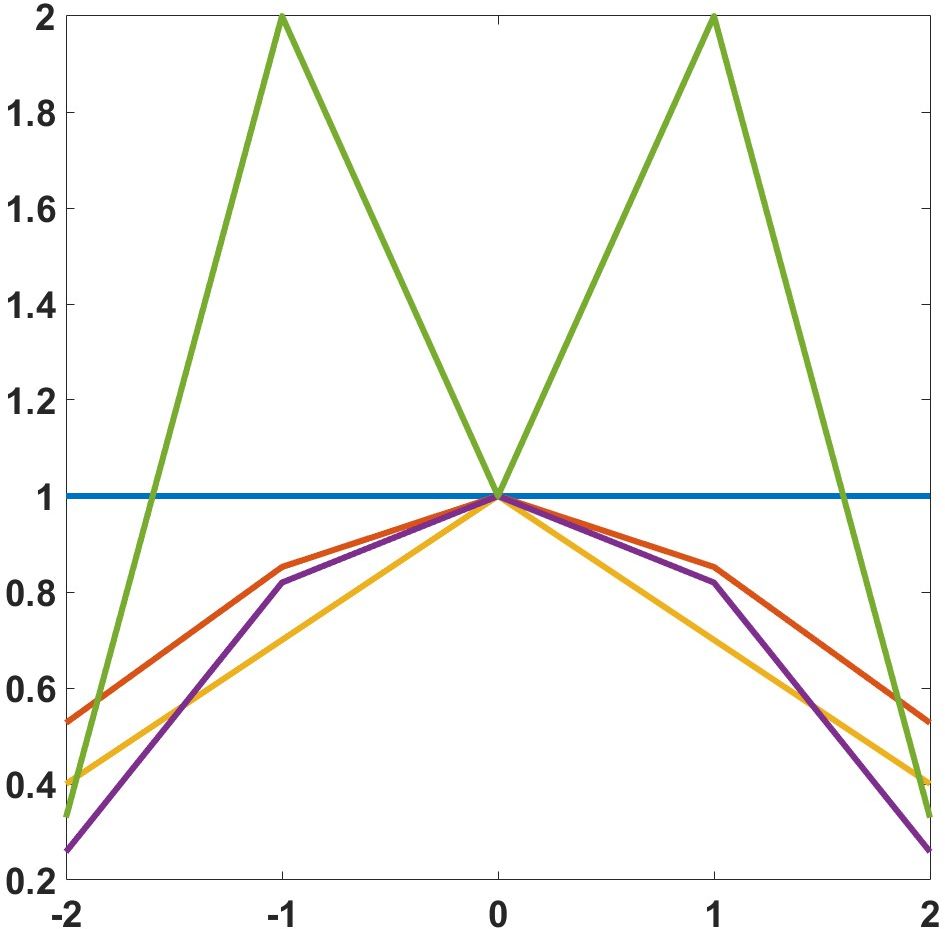} &
\includegraphics[width=0.40\textwidth]{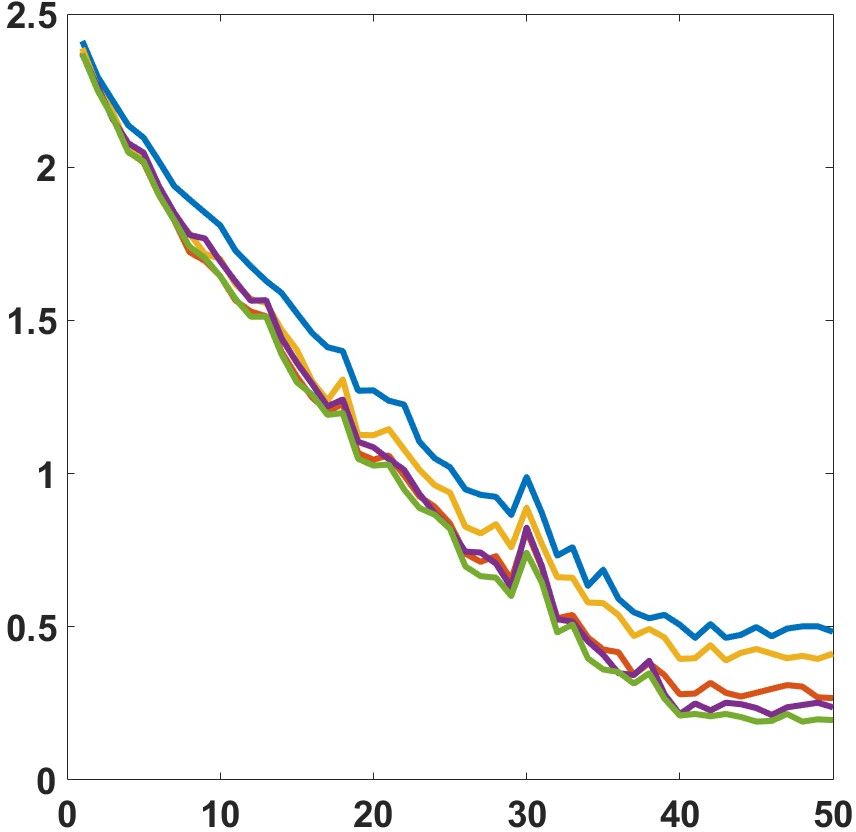} \\
(a) & (b) 
\end{tabular}
\caption{(a) Different density functions: uniform (blue); linear with slope equal to~$0.3$ (yellow); Gaussian with~$\sigma = 1.5$ (red); cubic (purple); our (green). All the density functions have the central value equal to~$1$. (b) Cross-entropy training loss ($y-$axis) for the different density functions within the ResNet network, 50 epochs ($x-$axis). \label{fig:starKernels}}
\end{figure}

\paragraph{Analysis on symmetry requirements of the density function}
In our tests, we have required a symmetric density function on both the kernel dimensions, i.e.,~$\bm{\alpha} = \bm{\beta}$ and~$\bm{\alpha}_i = \bm{\alpha}_{K - i + 1}, i = 1\ldots K$, with~$K$ size of the kernel. As a further verification of this property, we optimise the values of the density function, removing these two constraints. In particular, we define a first optimisation problem with a~$3 \times 3$ kernel and~$2$ optimisation variables,~$\bm{\alpha}_1$ and~$\bm{\alpha}_2$, with~$\bm{\alpha} = [\bm{\alpha}_1, 1, \bm{\alpha}_2]$, and~$\bm{\alpha} = \bm{\beta}$. Then, we define a second optimisation problem, with a~$3 \times 3$ kernel and~$2$ optimisation variables,~$\bm{\alpha}_1$ and~$\bm{\beta}_1$, with~$\bm{\alpha} = [\bm{\alpha}_1, 1, \bm{\alpha}_1]$ and~$\bm{\beta} = [\bm{\beta}_1, 1, \bm{\beta}_1]$. In both the cases, our optimisation model shows that the optimal values of the density function satisfy the symmetric property, i.e.,~$\bm{\alpha}_1 \simeq \bm{\alpha}_2$ and~$\bm{\alpha}_1 \simeq \bm{\beta}_1$ respectively.
\begin{table}[t]
\centering
\caption{Optimal value of~$\bm{\alpha}_1$ with a~$3 \times 3$ kernel size, with respect to the hyperparameters of the learning phase.\label{TAB:robustness1}}
\begin{tabular}{l|llllll}
\textbf{Stride} &1 &2 &4 &8 & &\\
\hline
$\bm{\alpha}_1$ & 0.42 & 0.42 & 0.47 & 0.49\\
\hline\hline
\noalign{\vskip 6pt}
\textbf{Epochs} &1 &2 &5 &10 &20 &50 \\
\hline
$\bm{\alpha}_1$ &1.48 &0.82 &0.53 &0.46 &0.42 &0.42 \\
\hline\hline
\noalign{\vskip 6pt}
\textbf{Number of images} & 10 & 30 & 60 & 120 \\ \hline
$\bm{\alpha}_1$ & 0.90 & 0.77  & 0.45 & 0.42\\
\hline\hline
\noalign{\vskip 6pt}
\textbf{Image size: (Rows, columns)} &$(32,32)$ &$(64,64)$ &$(128,128)$ &$(256,256)$ \\ \hline
$\bm{\alpha}_1$ & 0.31 & 0.31  & 0.42 & 0.42 \\
\hline \hline
\noalign{\vskip 6pt}
\textbf{Channels (c)}:~$L^1_{1c}$-$L^2_{cc}$-$L^3_{c1}$ & 1 & 2 & 4 & 8 & 16 \\ \hline
$\bm{\alpha}_1$ &0.66 & 0.44 & 0.42 & 0.42 & 0.42  
\end{tabular}
\end{table}
\paragraph{Analysis on density function robustness to learning hyperparameters}
The optimisation method computes the optimal values of~$\bm{\alpha}$ to minimise the objective function of Eq.~(\ref{eq:FUNCTIONAL}), whereas the learning model is applied to minimise the loss function of Eq.~(\ref{eq:wModel}) given a certain density function~(i.e., the iterations of Eq.~(\ref{eq:FUNCTIONAL})). We evaluate the robustness of the values of~$\bm{\alpha}$ to the hyperparameters of the learning model. In particular, Table~\ref{TAB:robustness1} shows the results of the optimal~$\bm{\alpha}_1$ value when optimising the density function of a~$3 \times 3$ kernel with different strides, epochs, data set dimension, image size, and learning model architecture (i.e., channels) values. In all the tests, the~$\bm{\alpha}_1$ value converges to the optimal value of~$0.42$. We underline that this value tends to increase when we reduce the complexity of the network (e.g., a reduced number of epochs or channels) and when we reduce the data information (e.g., the number of images in the data set, the stride value). For example, the~$\bm{\alpha}_1$ value passes from~$1.48$ to~$0.42$ when we increase the number of epochs from~$1$ to~$50$, and the~$\bm{\alpha}_1$ value passes from~$0.42$ to~$0.49$ when we increase the stride from~$1$ to~$8$. When we increase the redundancy of the learning phase (e.g., more images or epochs), the optimal density function increases the locality of the convolution in terms of feature extraction, thus reducing the relevance of the pixels close to the reference one. In contrast, reducing the image size (e.g., from~$256 \times 256$ to~$64 \times 64$) reduces the optimal value of~$\bm{\alpha}_1$ from~$0.42$ to~$0.31$, thus increasing the locality of the convolution. Finally, the number of channels affects the number of trainable parameters; also, in this case, the density function converges to the optimal value when increasing the complexity of the learning model.
\paragraph{Comparison with common density functions}
Our convolution with an optimal density function can be applied to any deep learning architecture. Given the \emph{deep residual learning} (ResNet) network~\cite{he2016deep}, we solve a multi-label classification problem with the STL-10 data set (500 images, 10 labels)~\cite{coates2011analysis} with different density functions applied to the convolution operator: uniform, Gaussian, linear, cubic, and our optimal density function (Fig.~\ref{fig:starKernels}a). During the training with 50 epochs, the different density functions have a similar cross-entropy loss (Fig.~\ref{fig:starKernels}b), while our optimal density function reaches the lowest value. Then, Table~\ref{tab:testloss} shows the loss value on the test data set with the different density functions. A uniform density function generates a cross-entropy loss value on the test data set of~$1.73$, which corresponds to a classification accuracy of around~$46\%$. Our optimal density function generates a loss value of~$1.54$ and a classification accuracy of~$53\%$. We underline that our goal is not the implementation of an efficient learning architecture, but the computation of the optimal density function for the convolution applied to a learning architecture, and the comparison with uniform and common density functions.
\subsection{Execution time\label{sec:EXPRES2}}
We discuss the execution time of the weighted convolution and the optimisation of the values of~$\bm{\alpha}$. Our tests are performed on a standard workstation with an AMD Ryzen 9 7845HX CPU with a 3GHz clock, 16 GB RAM, NVIDIA GeForce RTX 4070 GPU with 8GB vRAM.
\begin{table}[t]
\centering
\captionsetup{width=0.9\textwidth}
\caption{Concerning Fig.~\ref{fig:starKernels}(b), we report the loss function and the classification accuracy on the test data set of 30 images. Best result in bold.\label{tab:testloss}}
{\begin{tabular}{c|cccc|c}
Density function  & Uniform & Linear & Cubic & Gaussian & Our \\ \hline
Loss function & 1.73 & 1.70 & 1.65 & 1.68 &$\bm{1.54}$ \\
Classification accuracy &~$46\%$ &~$47\%$ &~$49\%$ &~$48\%$ &$\bm{53\%}$
\end{tabular}}
\end{table}
\paragraph{Convolution with density function}
We compare the execution time of the convolution with and without the application of the density function, where the non-application of the density function is equivalent in terms of numerical output to the application of a uniform density function with all the components equal to~$1$. Table~\ref{TAB:EXECUTION1} summarises the execution time of the convolution with and without density function for different learning model parameters (i.e., output channels and kernel size). Given a~$(2,3,512,512)$ image in the format \emph{batch size}, \emph{input channels}, \emph{rows}, and \emph{columns}, we test~$1$,~$3$, and~$6$ output channels with~$3\times 3$,~$5\times 5$, and~$7\times 7$ kernel size. The execution time of the convolution with the density function increases by around~$11\%$ with respect to the convolution without the density function. The execution time reduces when increasing the kernel size, as fewer patches are processed. Also, the execution time increases when increasing the output channels of the convolution. To test a single convolution operation, we define a~$2000 \times 2000$ image and a~$2000 \times 2000$ kernel size. Also in this case, the convolution with the density function increases the execution time by around~$11\%$ with respect to the convolution without the density function.
\begin{table}[t]
\centering
\captionsetup{width=0.9\textwidth}
\caption{Execution time in [ms] of the convolution with and without density function for different kernel sizes and output channels.\label{TAB:EXECUTION1}}
{\begin{tabular}{l|lll|lll|lll}
Output channels  & \multicolumn{3}{|c|}{1} &  \multicolumn{3}{c|}{3} &\multicolumn{3}{c}{6} \\ \hline
Kernel size &$3\times3$ &~$5\times5$ &~$7\times7$ &$3\times3$ &~$5\times5$ &~$7\times7$ &$3\times3$ &~$5\times5$ &~$7\times7$ \\ \hline
With density function  &  18    &   10   &    6  &   77   &   25    &   18   &   135   &    45   & 35 \\
Without density function  & 16     &  9  & 5   &   71   &   22   &   16    &  120    &   41    & 31
\end{tabular}}
\end{table}
\paragraph{Density function optimisation}
The computation of the optimal density function is computed through the \emph{DIRECT-L} optimisation method. The execution time depends on the number of variables (i.e., the size of the kernel) and the execution time of the learning model. Given a learning model with 200 images and 720 trainable parameters and trained for 20 epochs, Table~\ref{TAB:EXECUTION2} shows the execution time and the number of iterations of \emph{DIRECT-L} with different kernel sizes. The execution time passes from~$8,650$ seconds with~$3 \times 3$ kernel size to~$142,000$ seconds with~$7 \times 7$ kernel size.
\begin{table}[t]
\centering
\captionsetup{width=0.9\textwidth}
\caption{Execution time and the number of iterations of \emph{DIRECT-L} for different kernel sizes. The learning model is defined with~$20$ epochs,~$150$ images, stride~$s=1$, and output channels~$c = 8$.\label{TAB:EXECUTION2}}
{\begin{tabular}{c|ccc}
Kernel size &$3 \times 3$ &~$5 \times 5$  &~$7 \times 7$  \\ \hline
Execution time [s] & 8650 & 55,300 & 142,000 \\
Number of iterations & 260 & 840 & 995 
\end{tabular}}
\end{table}
\section{Conclusions and future work}\label{sec:CONCLUSION}
We have defined and computed the optimal density function for the weighted convolution operator in 2D images and learning problems. We have analysed the results with different kernel sizes and learning hyperparameters. The application of the optimal density function to the weighted convolution in learning models allows us to reduce the loss function of the network from~$12\%$ to~$53\%$ with respect to the standard convolution. As future work, we plan to apply our weighted convolution with the optimal density function to real-case deep learning problems (e.g., segmentation, classification) on large data sets and with state-of-the-art network architectures (e.g., \emph{efficientNet}~\cite{tan2020efficientnetrethinkingmodelscaling}, \emph{uNet}~\cite{ronneberger2015u}), with 2D and 3D images, and compare the accuracy of the learning when applying the standard and weighted convolution.

\textbf{Acknowledgements}
SC and GP are part of RAISE Innovation Ecosystem, funded by the European Union - NextGenerationEU and by the Ministry of University and Research (MUR), National Recovery and Resilience Plan (NRRP), Mission 4, Component 2, Investment 1.5, project “RAISE - Robotics and AI for Socio-economic Empowerment” (ECS00000035).


\bibliographystyle{alpha}
\bibliography{bibliography}

\clearpage
\begin{appendices}

\section*{Weighted convolution properties}\label{sec:INTROCT}
Recalling the definition of the weighted convolution as \mbox{$(f \ast g_{\varphi})(z) = \int_{\mathbb{R}} f(x) \cdot( g(z-x)  \varphi(z-x)) dx$} with~$f,g \in \mathcal{L}^1(\mathbb{R})$, we briefly discuss its main properties.

\textbf{Convolution theorem}
The Fourier transform of the weighted convolution satisfies the relation \mbox{$\mathcal{F}(f \ast g_{\varphi} )(v) = \mathcal{F}( f)(v) \cdot \mathcal{F}( g \cdot \varphi)(v)$}. In fact,
\begin{align*}
\mathcal{F}(f \ast g_{\varphi})(v) &= \int_{\Omega} (f \ast g_{\varphi})(z)  e^{-2 \pi i z \cdot v} dz && \\
&= \int_{\Omega\times\Omega}  f(x) \cdot ( g(z-x) \varphi(z-x) dx) \cdot e^{-2 \pi i z \cdot v} dz && \\
&= \int_{\Omega} f(x)  \left( \int_{\Omega} g(y) \varphi(y)   \cdot e^{-2 \pi i (x+y) \cdot v} dy\right) dx && y = z - x \\
&= \int_{\Omega} f(x) \cdot e^{-2 \pi i x \cdot v}  \left( \int_{\Omega} g(y) \varphi(y)  \cdot e^{-2 \pi i y \cdot v} dy \right) dx && \\
&= \underbrace{\int_{\Omega} f(x) \cdot e^{-2 \pi i x \cdot v} dx}_{\mathcal{F}( f)(v) }  \underbrace{\int_{\Omega} g(y) \varphi(y)   \cdot e^{-2 \pi i y \cdot v} dy.}_{\mathcal{F}( g \cdot \varphi)(v)}  &&\\
\end{align*}
\textbf{Commutativity}
\begin{align*}
(f \ast g_{\varphi} )(z) &= \int_{-\infty}^{\infty} f(x) (g(z-x) \varphi(z-x)) dx && \\
&= -\int_{\infty}^{-\infty} f(z-y) ( g(y) \varphi(y)) dy && y = z-x \\
&= \int_{-\infty}^{\infty} f(z-y) ( g(y) \varphi(y)) dy && \\
&= (g_{\varphi} \ast f)(z) &&
\end{align*}
\textbf{Differentiability}
Computing the derivative of
$$h(z) = (g_{\varphi} \ast f)(z) = \int_{\Omega} f(z-x) \cdot ( g(x) \varphi(x)) dx,$$
with respect to the filter~$g$, we get that
\begin{align*}
\frac{\partial h(z)}{\partial g(s)}  &= \frac{\partial}{\partial g(s)}  \int_{\Omega} f(z-x) \cdot ( g(x) \varphi(x)) dx \\
&= \frac{\partial}{\partial g(s)} f(z-s) \cdot ( g(s) \varphi(s)) \\
&= f(z-s) \cdot \varphi(s) .
\end{align*}
\textbf{Density functions identity}
The density function applied to the input signal~$f$ equals the shifted density function applied to the filter~$g$. Given two density functions~$\varphi$ and~$\psi$, we solve~$ (f \ast g_\varphi)(z) = (f_\psi \ast g)(z)$, i.e.,
$$ \int_{\Omega} f(x) \cdot (g(z-x) \cdot \varphi(z-x)) dx =  \int_{\Omega} (f(x) \cdot \psi(x)) \cdot g(z-x)  dx.$$
Imposing
$$ \int_{\Omega} f(x) \cdot g(z-x) \cdot ( \varphi(z-x) - \psi(x) ) dx = 0,$$
we get that~$\varphi(z-x) = \psi(x)$.\\

\textbf{Young's inequality}
Assuming \mbox{$\varphi\in\mathcal{L}^{\infty}(\mathbb{R})\cap\mathcal{L}^{1}(\mathbb{R})$}, the weighted convolution satisfies the relation $f \ast g_{\varphi} \in \mathcal{L}^{1}(\mathbb{R})$. In fact,
\begin{align*}
\left \|(f \ast g_{\varphi})(z) \right \|_1 &= \int_{\mathbb{R}} \left | \int_{\mathbb{R}} f(x)   (g(z-x)   \varphi(z-x)) dx \right  | dz && \\
 &\leq \int_{\mathbb{R}} \int_{\mathbb{R}}  \left | f(x) \right |    \left | g(z-x) \right |    \left | \varphi (z-x) \right | dx dz && \varphi \in \mathcal{L}^1(\mathbb{R}) \\
 &\leq \int_{\mathbb{R}}\left | f(x) \right |  \int_{\mathbb{R}}   \left | g(z-x) \right |    \left | \varphi (z-x) \right | dz dx && \\
 &\leq \left \|f\right \|_1 \int_{\mathbb{R}} \left | g(z-x) \right |    \left | \varphi (z-x) \right | dz && \\
&\leq \left \|f\right \|_1 \int_{\mathbb{R}} \left | g(y) \right |    \left | \varphi (y) \right | dy && y = z - x\\
&\leq \|\varphi\|_{\infty} \left \|f\right \|_1 \int_{\mathbb{R}} \left | g(y) \right |   dy &&\\
&\leq \|\varphi\|_{\infty}   \left \|f \right \|_1   \left \|g \right \|_1   && \\
&< \infty &&
\end{align*}
%
%
\end{appendices}
\end{document}